%% file: paper.tex
\documentclass[letterpaper, 10 pt, conference]{ieeeconf}  % Comment this line out if you need a4paper

\IEEEoverridecommandlockouts                              % This command is only needed if 
\usepackage{bm,amsmath,amsfonts,amssymb,dsfont} % math
\usepackage{multirow} % for table
\usepackage{graphicx} % for figures
\graphicspath{{images/}}
\usepackage{hyperref} % for URL
\hypersetup{
    colorlinks=true,
    linkcolor=blue,
    filecolor=magenta,
    urlcolor=blue,
}

\newcommand{\NewR}{\ensuremath{\mathds{R}}}
\newcommand{\mat}[1]{{\ensuremath{{\mathbf{#1}}}}}

\title{\LARGE \bf
Investigations on Output Parameterizations of Neural Networks for Single Shot 6D Object Pose Estimation
}
\author{
Kilian Kleeberger$^{1}$,
Markus Völk$^{1}$,
Richard Bormann$^{1}$,
and
Marco F. Huber$^{1,2}$

\thanks{$^{1}$Fraunhofer Institute for Manufacturing Engineering and Automation IPA,
        Nobelstra{\ss}e~12, 70569 Stuttgart, Germany
        {\tt\small kilian.kleeberger@ipa.fraunhofer.de}}%
\thanks{$^{2}$Institute of Industrial Manufacturing and Management IFF, University of Stuttgart,
        Allmandring~35, 70569 Stuttgart, Germany
        {\tt\small marco.huber@ieee.org}}%
}
\begin{document}

\maketitle
\thispagestyle{empty}
\pagestyle{empty}

%%%%%%%%%%%%%%%%%%%%%%%%%%%%%%%%%%%%%%%%%%%%%%%%%%%%%%%%%%%%%%%%%%%%%%%%%%%%%%%%
\begin{abstract}
Single shot approaches have demonstrated tremendous success on various computer vision tasks. Finding good parameterizations for 6D object pose estimation remains an open challenge. In this work, we propose different novel parameterizations for the output of the neural network for single shot 6D object pose estimation. Our learning-based approach achieves state-of-the-art performance on two public benchmark datasets. Furthermore, we demonstrate that the pose estimates can be used for real-world robotic grasping tasks without additional ICP refinement.
\end{abstract}
%/elegant/efficient
%or instance segmentation
%is not trivial /
%, such as image classification, segmentation, object detection, etc.

%%%%%%%%%%%%%%%%%%%%%%%%%%%%%%%%%%%%%%%%%%%%%%%%%%%%%%%%%%%%%%%%%%%%%%%%%%%%%%%%
\section{Introduction}

\begin{figure}
\centering
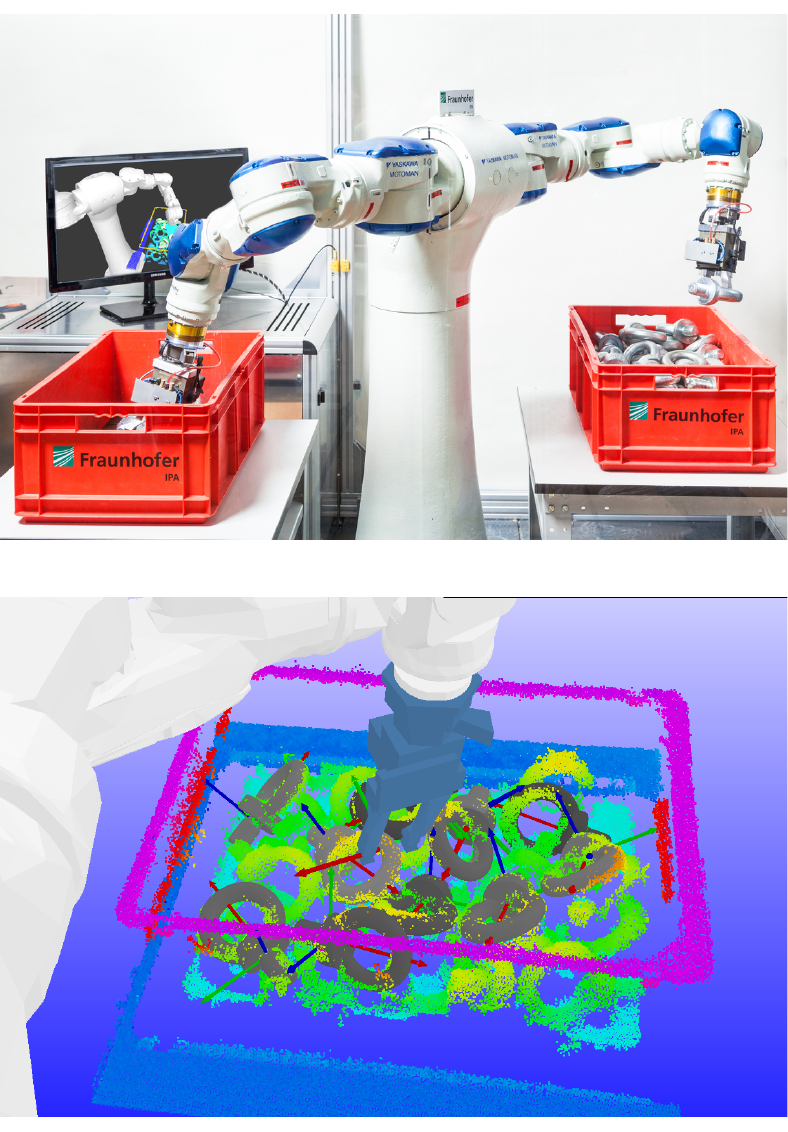
\caption{
    (a) Real-world robot cell for random bin-picking.
    (b) 3D point cloud (colored) with pose estimates (grey) of \mbox{OP-Net AP} on real-world data without ICP refinement for ring screws. The gripper (blue) is visualized at the chosen grasp pose for execution together with the joint configuration of the robot (white).
}
\label{fig:real_world_robot_cell_and_pose_estimates}
\end{figure}

% Importance OPE:
%((Pose estimation is important for numerous application fields such as robotic grasping and manipulation, augmented reality, and autonomous driving.))
%Robust/
A reliable, fast, and accurate method for 6D object pose estimation is a crucial prerequisite for many robotic grasping and manipulation tasks.
Based on a single depth image of known rigid objects, our goal
%for robust pose estimation based on single depth images for know rigid objects.
%The goal
is to estimate the translation vector $\mat{t} \in \NewR^3$ and rotation matrix ${\mat R \in \mathrm{SO}(3)}$ of an object relative to
%a given reference frame, e.g.,
the sensor coordinate system. Using this information,
%The task of the robot is to localize the objects and then
a robot can plan a kinematically feasible and collision-free path
%towards the object
to pick and place the object as visualized in Fig.~\ref{fig:real_world_robot_cell_and_pose_estimates}.

% Challenges OPE:
%The task
Object pose estimation is challenging because of a potentially very high amount of clutter and occlusion in the scene. Object symmetries propose different annotations for identical observations and typically have to be addressed explicitly for learning-based approaches~\cite{OP_Net,PPR_Net,CosyPose}. Furthermore, occlusions can also lead to pose ambiguities and different lighting conditions affect the appearance of the object in the image. The task is also challenging due to missing, wrong, and noisy depth information.

% problems:
% - noise
% - wrong
% - missing/incomplete

\begin{figure*}[t]
\centering
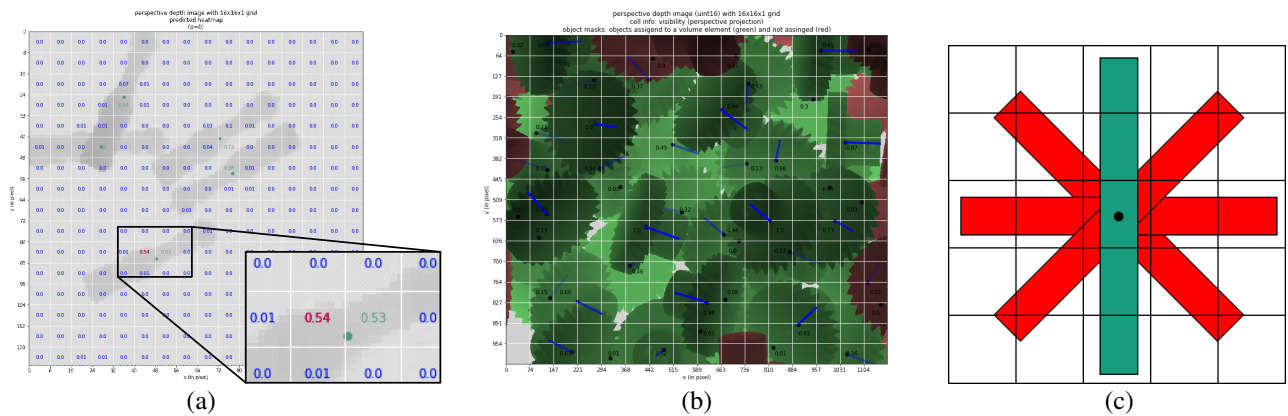
\caption{
Major limitations of
%discretization-based % add?
single shot approaches to 6D object pose estimation using an assignment based on the origin of the object:
%(\mbox{OP-Net}, ...):
(a) Exemplary prediction for the presence of an object origin for the pepper object from the Sil\'{e}ane dataset~\cite{Bregier.2017}:
%(p channel):
If an object origin is close to the border of a volume element, it might also be incorrectly detected by the neighboring spatial location (red), potentially with a higher confidence. Both cells detect the object with low confidence ($\hat{p}\in[0,1]$).
(b) Exemplary ground truth sample for the gear object from the Sil\'{e}ane dataset~\cite{Bregier.2017}: Objects, which are assigned to the ground truth tensor, are masked in green. Some objects at the border of the image are well
%/nicely
visible but cannot be predicted because their object origin
%of the object
is outside of the image (masked in red).
(c) The pose of well
%/nicely
visible objects (red) cannot be predicted because each spatial location can only predict one pose. The object with the highest visibility (green) is used in the ground truth tensor.
}
\label{fig:limitations}
\end{figure*}

% Power of learning based methods:
Learning-based methods have demonstrated tremendous success on various benchmarks~\cite{OPE_Challenge_Bin_Picking_IROS,BOP_Challenge_2020}.
%proven to be ...
%BOP Challenge 2020~\cite{BOP_Challenge_2020}
Due to the high amount of data needed for training deep neural networks,
%Especially for learning-based methods,
using simulations is an attractive choice as they provide an abundant source of data with flawless annotations~\cite{Kleeberger_Review_Article,Fraunhofer_IPA_Bin_Picking_dataset,BlenderProc}.
%
% Advantages of single shot approaches:
Single shot approaches are very fast because they require a single forward pass of the neural network and consider global scene context instead of looking at local image patches only.
They have demonstrated tremendous advantages in speed, also for instance segmentation~\cite{YOLACT,YOLACTupgrade}.
%single shot approaches to object detection have outperformed multi-stage approaches to object detection such as (variants of R-CNN).
%The same also holds for 6D object pose estimation~\cite{OP_Net,PPR_Net}.

% OP-Net is cool:
\mbox{OP-Net}~(Object Pose Network)~\cite{OP_Net} outperformed PPR-Net~\cite{PPR_Net}, the winning method of the ``Object Pose Estimation Challenge for Bin-Picking'' at IROS 2019~\cite{OPE_Challenge_Bin_Picking_IROS}, on the Sil\'{e}ane dataset~\cite{Bregier.2017}
based on the
% evaluation metric
commonly used evaluation metric for 6D object pose estimation from Br\'{e}gier et al.~\cite{Bregier_Pose_Distance,Bregier.2017}, which considers all possible kinds of object symmetries and can handle scenarios in bulk. The metric requires the recovery of the pose of all objects in the image
%/scene
with a visibility of 50\% or higher.
% the metric has bin used in a robotic challenge for object pose estimation~\cite{OPE_Challenge_Bin_Picking_IROS} and various scientific publications~\cite{OP_Net,PPR_Net} [xx].
%
Furthermore, \mbox{OP-Net} has been extended to model-based robotic pick-and-place~\cite{PQ_Net}.

Despite providing robust pose estimates at a very high speed,
%Still,
the method comes with shortcomings regarding the parameterization of the output which limit the performance. Several limitations are visualized in Fig.~\ref{fig:limitations}~(see Section~\ref{sec:problem_statement} for more details).
% Simple for semantic segmentation
While the design of the network output is trivial for tasks such as semantic segmentation (one-hot encoded class vector for each pixel), finding a suitable output is challenging for tasks such as object detection~\cite{YOLOv1,YOLOv2,SSD}, object pose estimation~\cite{OP_Net,BB8,SS_OPE_CVPR_2018}, or instance segmentation~\cite{MaskRCNN,YOLACT,YOLACTupgrade}. In this work, we
%aim to
address these challenges and give solutions by proposing different novel parameterizations for the output of the neural network. Fig.~\ref{fig:real_world_robot_cell_and_pose_estimates}~(b) shows pose estimates of our approach on real-world data, which are accurate enough for reliable grasping with a robot even without ICP refinement.
%without ICP refinement which are accurate enough for
%robust/
%reliable grasping with a robot.
Videos of our experiments are available at \url{https://owncloud.fraunhofer.de/index.php/s/6QsRj5sSty7fI9c}.

% Contributions:
In summary, the main contributions of this work are:
\begin{itemize}
    \item Propose different novel parameterizations of the output of the neural network which solve problems of current methods using an origin-based assignment
    \item Extensive evaluation on public benchmark datasets
    \item Real-world robot demonstration of the approach for bin-picking
\end{itemize}

%%%%%%%%%%%%%%%%%%%%%%%%%%%%%%%%%%%%%%%%%%%%%%%%%%%%%%%%%%%%%%%%%%%%%%%%%%%%%%%%
\section{Related Work}
\label{sec:related_work}

% benchmarks / challenges:

% Occluded Object Challenge~\cite{Occluded_Object_Challenge}

% Object pose estimation is a long-standing challenge and an open field of research since the early days of computer vision.
Object pose estimation is
a fundamental task
%one of the oldest problems
in computer vision and an active field of research~\cite{CosyPose}.
Challenges such as the ``Object Pose Estimation Challenge for Bin-Picking''~\cite{OPE_Challenge_Bin_Picking_IROS}, SIXD Challenge~\cite{SIXD_Challenge_2017}, or BOP (Benchmark for 6D Object Pose Estimation) Challenge~\cite{Internet_BOP_Challenge,BOP_Challenge_2019,BOP_Challenge_2020,Paper_BOP} aim to capture and advance the state of the art in the field. Contrary to the SIXD Challenge~\cite{SIXD_Challenge_2017} in 2017 and the BOP Challenge in 2019~\cite{BOP_Challenge_2019}, in the BOP Challenge 2020~\cite{BOP_Challenge_2020} learning-based approaches caught up with classical approaches, including methods based on point pair features~\cite{PPF}, which dominated previous editions of the challenge.
%
%
%
% Classical methods:
%
% CNN-based methods:
CNN-based methods gained popularity in recent years, are a dominant research direction, and outperformed classical methods~\cite{BOP_Challenge_2020} based on feature~\cite{PPF} or template~\cite{Hinterstoisser_ADD/ADI_LINEMOD+} matching.

% multi-stage approaches:
% multi-stage

% single shot approaches:
% single-stage regression approaches

PPR-Net~\cite{PPR_Net}, the winning method of the ``Object Pose Estimation Challenge for Bin-Picking'' at IROS 2019~\cite{OPE_Challenge_Bin_Picking_IROS} uses PointNet++~\cite{PointNet++}, estimates a 6D pose for each point in the point cloud and gets the final pose hypothesis
%after/
by averaging the resulting pose clusters in 6D space.
% \mbox{OP-Net}~\cite{OP_Net} outperforms... --> chapter 1
\mbox{OP-Net}~\cite{OP_Net} introduces a spatial discretization of the measurement volume of the sensor and solves a regression task for each resulting volume element.
Other single shot approaches to object pose estimation output the 2D projections of the 3D bounding box and use a P$n$P algorithm~\cite{PnP} to compute
%/get/determine
the 6D pose~\cite{BB8,SS_OPE_CVPR_2018,PVNet,DOPE}. This results in a suboptimal two-stage process~\cite{SS_OPE_CVPR}. Our single shot variants directly regress the full 6D pose.
% Contrary to \mbox{OP-Net}, these methods requires an additional step (P$n$P).
The methods provide
%/estimatef
the pose of multiple objects simultaneously (single shot).
%
% Limitations pixel/point-based approaches:
Approaches with pixel- or point-wise predictions~\cite{PPR_Net}
%[xx]
%ToDo: Limitations PPR-Net
are slower because of
%not using a compact
using a less compact parameterization and using post processing for averaging the predicted poses.
%(slow; one pose per pixel and PP for averaging)

% other pixel/point based regression methodS?!

% less expensive post-processing
% aim for keeping a compact parameterization of the output --> good runtime

\section{Problem Statement}
\label{sec:problem_statement}
%In this work, our goal is to estimate the 6D pose of known rigid objects from a single category relative to the sensor coordinate system based on a single depth image.
Based on a single depth image, our goal is to estimate the 6D pose of known rigid objects from a single category relative to the sensor coordinate system.
Current single shot systems for 6D object pose estimation~\cite{OP_Net,SS_OPE_CVPR_2018,PQ_Net} discretize the measurement volume of the sensor in $S_x \times S_y$ volume elements.
%%%%%
% KBK-ET: Schiefer (Rechtecks-)Pyramidenstumpf
% Pyramidenstumpf: truncated pyramid
% Rechtecks-Pyramide: rectangular pyramid
% Schiefe Pyramide: oblique pyramid; "An oblique pyramid is one where the apex is not over the center of the base." Source: https://www.mathopenref.com/pyramidoblique.html#:~:text=An%20oblique%20pyramid%20is%20one%20where%20the%20apex,to%20calculate%20its%20surface%20area%20no%20longer%20work.
% --> truncated oblique rectangular pyramid
%%%%%
% rectangular and not quadratic because res_x!=res_y and/or S_x!=S_y
%The shape of these volume elements in perspective projection are truncated oblique rectangular pyramids.
Due to the perspective projection, the shape of these volume elements corresponds to truncated oblique rectangular pyramids.
At each spatial location, a feature vector of the object is assigned, which results in a 3D output tensor. The feature vector \mbox{$\mat{x}=[p,v,x,y,z,\varphi_1,\varphi_2,\varphi_3]$} comprises the presence $p=1$
% \in \{0,1\}$
of an object, the visibility $v \in [0,1]$, the positions $x$, $y$, $z$, and the Euler angles $\varphi_1$, $\varphi_2$, $\varphi_3$. For objects with a revolution symmetry,
it is sufficient to predict the angles $\varphi_1$ and $\varphi_2$, as
%predicting the angles $\varphi_1$ and $\varphi_2$ only is sufficient, because
the rotation around the $z$-axis with $\varphi_3$ results in identical observations.
% or a, b only for ... symmetric objects.
In case multiple objects fall into the same volume element, the object with highest visibility is chosen for assignment. For all spatial locations without an object, a zero vector is assigned.

% Limitations discretization-based approaches (\mbox{OP-Net}, ...):
The parameterization of these single shot systems for 6D object pose estimation~\cite{OP_Net,SS_OPE_CVPR_2018,PQ_Net}
%(\mbox{OP-Net}, 3D bb and P$n$P, ...)
%face/
comes with several limitations,
% Limitations current system (\mbox{OP-Net}~\cite{OP_Net})
%Still, the approaches come with several limitations.
% as discussed in~\cite{OP_Net} in chapter XX. --> mention in this paper!
% we identified 3 mayor limitations as
which are presented in Fig.~\ref{fig:limitations}.
% Limitation 1:
In case the origin of an object is close to the border of a volume element, the object might also be detected by the neighboring volume element (also with higher confidence). This results in duplicates and the neighboring cell has not been trained to output useful pose information. In Fig.~\ref{fig:limitations}~(a), both cells detect the object with low confidence.
% Limitation 2:
Furthermore, the single shot approaches to object pose estimation
% (such as \mbox{OP-Net} or Tekin et al.~\cite{SS_OPE_CVPR_2018})
cannot predict objects with their
%reference point
object origin outside of the image (see objects masked in red in Fig.~\ref{fig:limitations}~(b)).
% Limitation 3:
Moreover, the parameterization is not unique when multiple objects fall into the same spatial location with their object origin (see Fig.~\ref{fig:limitations}~(c)). The objects are neglected in the ground truth and therefore cannot be predicted at test time.
In this
%paper/
work, we
%aim to
%tackle/
address these
%(challenges/)
shortcomings mentioned above by providing different novel parameterizations of the output to overcome the limitations of previous works and improve the performance of single shot approaches.

%%%%%%%%%%%%%%%%%%%%%%%%%%%%%%%%%%%%%%%%%%%%%%%%%%%%%%%%%%%%%%%%%%%%%%%%%%%%%%%%
\section{Approach}

This section describes
%the general
%approach/
%concept,
various novel output parameterizations of the neural network, the loss function, the averaging of poses for variants with multiple predictions per object, and the technique for a robust \mbox{sim-to-real} transfer.

%%%%%%%%%%%%%%%%%%%%%%%%%%%%%%%%%%%%%%%%%%%%%%%%%%%%%%%%%%%%%%%%%%%%%%%%%%%%%%%%
%\subsection{Concept}

%%%%%%%%%%%%%%%%%%%%%%%%%%%%%%%%%%%%%%%%%%%%%%%%%%%%%%%%%%%%%%%%%%%%%%%%%%%%%%%%
\subsection{Output Parameterizations}
In this section, we describe different parameterizations
%/variants
of the neural network output.

%%%%%%%%%%%%%%%%%%%%%%%%%%%%%%%%%%%%%%%%%%%%%%%%%%%%%%%%%%%%%%%%%%%%%%%%%%%%%%%%
%\subsubsection{Original version (OP-Net)}

%%%%%%%%%%%%%%%%%%%%%%%%%%%%%%%%%%%%%%%%%%%%%%%%%%%%%%%%%%%%%%%%%%%%%%%%%%%%%%%%
\iffalse
\subsubsection{Automatic Discretization (OP-Net AD)}

Original version (\mbox{OP-Net}) with automatic grid configuration
automatically determine the parameters $S_x$ and $S_y$
automatic discretization (AD)

to not miss objects with $v \geq 0.5$.

$S_z=1$

iterate over the whole training dataset (brute force)

As the dimension of the output is change, also the dimension of the input varies. In our experiments, we ensure that is input resolution is greater or equal to 128, i.e., that this variant does not observe a less detailed depth image.

The goal is to miss less object by the parameterization. Still,
%this variant does not solve the issue visualized in Fig. XX X and X.
the issues visualized in Fig.~\ref{fig:limitations} and (b) and (c) may still occur, with ... (other object more dominant) less frequent.
\fi

%%%%%%%%%%%%%%%%%%%%%%%%%%%%%%%%%%%%%%%%%%%%%%%%%%%%%%%%%%%%%%%%%%%%%%%%%%%%%%%%
\subsubsection{Extended Volume Elements (OP-Net EVE)}
For this variant, we also assign the feature vector $\mat{x}$
%of the object
to the neighboring volume elements if the origin of the object coordinate system is close to the border of a volume element
%as visualized in Fig.~\ref{fig:limitations}~(a)
%This extension is done to cells to the left/right and top/bottom as well as for combinations (i.e., diagonal).
% why?! motivation
because the object might also get incorrectly localized by the neighboring spatial location
%when it is close to the border of a volume element
as visualized in Fig.~\ref{fig:limitations}~(a).
%we also assign the ground truth feature vector to the neighboring spatial location.
%To address this issue, we train the network to predict correct features, for pose estimates which are close to the border and are therefore detected by the wrong spatial location.
% loss signal jumps during training --> ~0.5 at test time
With this modification, the model does not get a jumping loss for challenging border cases during training and the accepted duplicate pose estimates can be averaged (see Section~\ref{sec:pose_averaging}).

\begin{figure}
\centering
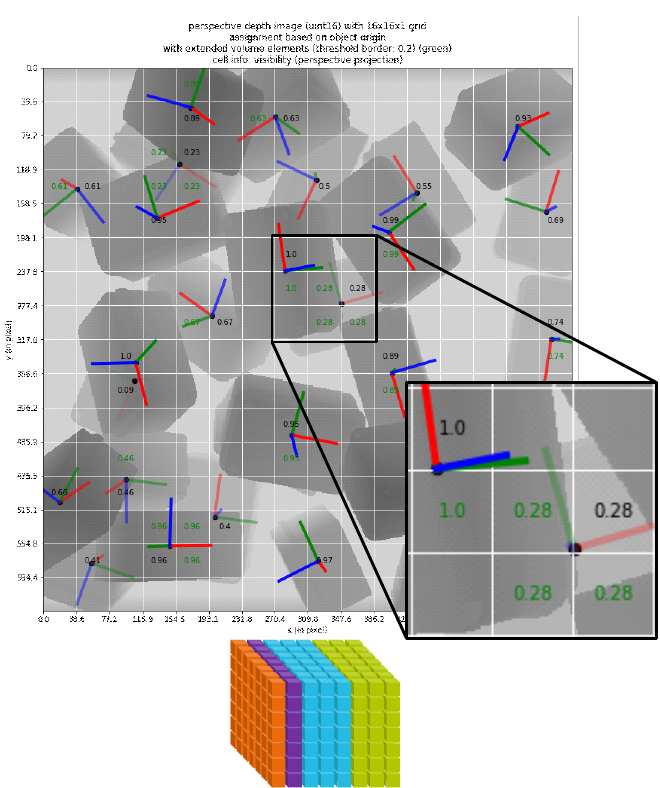
\caption{
(top) Exemplary ground truth sample for extended volume elements (EVE) for the T-Less 29 object from the Sil\'{e}ane dataset~\cite{Bregier.2017}. The feature vector of an object is also assigned to the neighboring elements if the origin of the object is close to the border of the cell. The extended elements are highlighted in green. The cell information indicates the visibility of the object.
(bottom) Resulting output tensor (shape unchanged).
}
\label{fig:EVE}
\end{figure}

% overwrite other origins?! No...
With these extensions to neighboring volume elements, we do not overwrite origins of other objects. In case multiple objects extend to the same volume element, we assign the feature vector of the object with highest visibility.
% threshold=0.2
In our experiments, we use a distance threshold from the border of 0.2 of the object origin relative to the spatial location. In the original version, the neighboring volume elements are not trained to propose correct pose
%/feature
%estimates
information. Fig.~\ref{fig:EVE} visualizes an exemplary ground truth sample
%from the Sil\'{e}ane dataset~\cite{Bregier.2017}
with extended volume elements.

% estimate the position relative to the image --> section "loss function"
% ...to also get ... for the limitation visualized in Fig.~\ref{fig:limitations}~(a).

%%%%%%%%%%%%%%%%%%%%%%%%%%%%%%%%%%%%%%%%%%%%%%%%%%%%%%%%%%%%%%%%%%%%%%%%%%%%%%%%
\subsubsection{Additional Points (OP-Net AP)}

\begin{figure*}[htb]
\centering
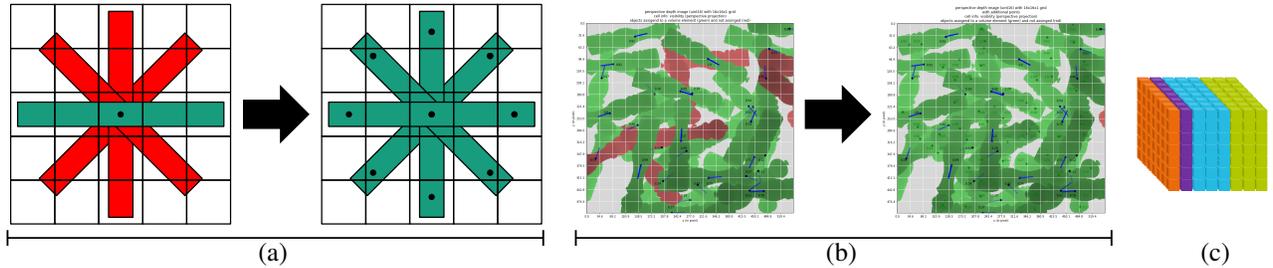
\caption{
(a) Concept for additional points (AP): Objects masked in green are assigned to a volume element
%to the output tensor
and objects masked in red are missed. With additional points much fewer objects are missed by the parameterization and, therefore, also at test time.
(b) An exemplary ground truth sample for the pepper object from the Sil\'{e}ane dataset~\cite{Bregier.2017} with two additional points. (c) Resulting output tensor (shape unchanged).
}
\label{fig:AP}
\end{figure*}

% Define additional points
An object may be well
%/nicely
visible in the image (i.e., high visibility), but another more dominant object (higher visibility) is already assigned to the volume element as visualized in Fig.~\ref{fig:limitations}~(c). As the spatial locations in the surrounding of the object do not have the task to predict a pose, they can be used for estimating additional ground truth information based on additional reference points on the object which are defined relative to the object coordinate system. Objects which are visible at the border of an image but with their origin outside can be predicted by this variant, because the additional points are inside the image (contrary to the object origin itself). Therefore, this variant addresses the limitation in Fig.~\ref{fig:limitations}~(b) and (c) plus implicitly also (a) because it is very unlikely that the object origin and all additional points are very close to a border. Fig.~\ref{fig:AP} visualizes the concept and demonstrates that much fewer objects will be missed by the used metric for objects pose estimation.

% Do not overwrite main points?
With additional points, we do not overwrite origins of other objects and assign the feature vector of the object with highest visibility for conflicting spatial locations.
% Requirements?
For objects with a revolution symmetry (e.g., candlestick, pepper, gear, gear shaft), the additional
%(reference)
points have to lie on the axis of symmetry.
For cyclic symmetries (e.g., brick, tless 20, tless 29, ring screw), the additional points have to permute
%change
the location
%be at the same location
when applying the steps of rotation.

% estimate the position relative to the image --> section "loss function"

%%%%%%%%%%%%%%%%%%%%%%%%%%%%%%%%%%%%%%%%%%%%%%%%%%%%%%%%%%%%%%%%%%%%%%%%%%%%%%%%
%\subsubsection{3D discretization}
%\subsubsection{Discretization in $z$-Direction (Z)}
\subsubsection{Discretization in z-Direction (OP-Net Z)}
\begin{figure}
\centering
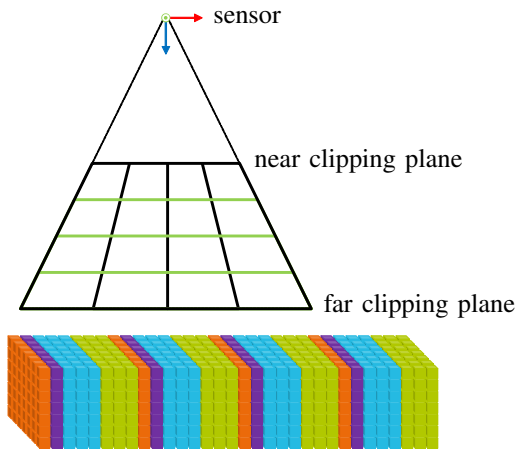
\caption{
(top)
%Exemplary ground truth sample for discretization in $z$-direction:
Discretization of the measurement volume of the 3D sensor in $z$-direction in addition to the $x$- and $y$-direction. The green lines indicate the introduced discretization in $z$-direction.
(bottom) Resulting output tensor ($S_z$ times the vanilla \mbox{OP-Net} tensor).
}
\label{fig:Z}
\end{figure}

In addition to the already used spatial discretization in $x$- and $y$-direction, we introduce a variant with additional discretization in $z$-direction from the near to the far clipping plane of the sensor. Fig.~\ref{fig:Z} visualizes this concept for the perspective projection. With this, we address the limitation in Fig.~\ref{fig:limitations}~(c) because the objects lie at different heights. This parameterization misses much fewer objects, which are relevant for the evaluation metric, i.e., objects with visibility $v \geq 0.5$. With the 8-dimensional feature vector, the output tensor is of shape $S_x \times S_y \times 8 \cdot S_z$ for this variant. In our experiments, we choose $S_x=S_y=S_z=16$.
%((The number of outputs of the neural network is highly increased. Our experiments demonstrate that this generally does not hurt the performance.))

%%%%%%%%%%%%%%%%%%%%%%%%%%%%%%%%%%%%%%%%%%%%%%%%%%%%%%%%%%%%%%%%%%%%%%%%%%%%%%%%
\subsubsection{Multiple Poses (OP-Net MP)}
% predict multiple poses per volume element / spatial location
As the vanilla \mbox{OP-Net} can predict one 6D pose per volume element only, we propose a variant that outputs multiple poses per spatial location to be able to predict the pose of less dominant objects in terms of visibility.
This addresses the limitation
%of the approach as
visualized in Fig.~\ref{fig:limitations}~(c).
%, we also introduce a variant which predicts multiple object poses per spatial location.
Each spatial location predicts $P$ feature vectors.
% S=8 (8x8 grid) and P=3
For our experiments, we use $S_x=S_y=8$, $S_z=1$ (i.e., no discretization in $z$-direction), and $P=3$. For practical scenarios, e.g., random bin-picking, it is very unlikely that more than three objects with $v \geq 0.5$ fall into the same volume element
%with their origin
with the chosen configuration (this does not happen in the datasets).

\begin{figure}
\centering
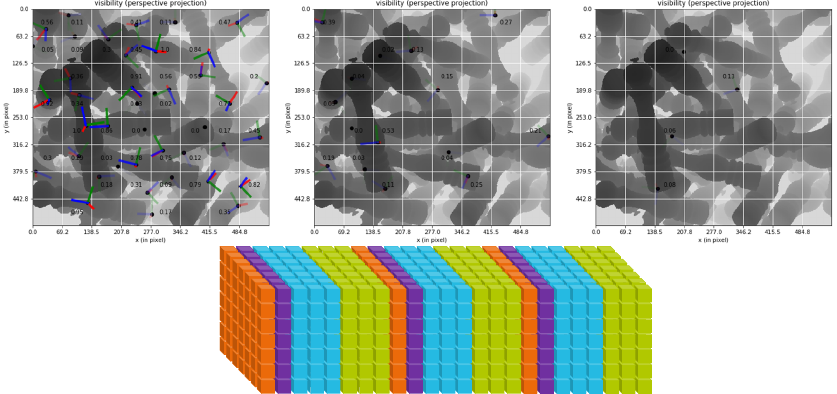
\caption{
(top) Exemplary ground truth sample for predicting multiple poses (MP) per spatial location for the pepper object from the Sil\'{e}ane dataset~\cite{Bregier.2017}.
(bottom) Resulting output tensor ($P$ times the vanilla \mbox{OP-Net} tensor).
}
\label{fig:MP}
\end{figure}

We condition the additional output feature maps for the presence of an object on the channel, i.e., each channel predicts whether the next channel comprises further objects.
%2nd or 3rd object, i.e., that the model does not need to re-predict the whole channel for the presence of an object.
The objects are ranked and assigned to the individual channels based on their visibility.
%assignment criterion (highest visibility in our case).
Fig.~\ref{fig:MP} shows an exemplary ground truth sample for predicting multiple poses per volume element.

%%%%%%%%%%%%%%%%%%%%%%%%%%%%%%%%%%%%%%%%%%%%%%%%%%%%%%%%%%%%%%%%%%%%%%%%%%%%%%%%
%\subsubsection{Assignment based on Resized Segmentation Image (OP-Net SI)}
\subsubsection{Resized Segmentation Image (OP-Net SI)}
Instead of assigning the objects to the ground truth tensor based on the origin, we propose to use the segmentation masks.
%and assign...
We resize the segmentation image (comprises the IDs of the objects) to \mbox{$32 \times 32$} pixel and assign the features of the objects to the tensor
%whose ID the image comprises.
based on the ID of the pixels.
% similar to PPR-Net~\cite{PPR_Net}
% our parameterization is more compact and fast for pp because less predictions that need to be clustered and averaged are made.
This variant addresses the limitations in Fig.~\ref{fig:limitations}~(a), (b), and (c).
% In our experiments, we use $S_x=S_y=32$ ($S_z=1$ and $P=1$).
%This variant also provides a coarse instance segmentation which can be used for ICP refinement.
%
% Due to
Because of the dense parameterization, we use a linear weighting for the pose error, i.e., $\lambda_3=v$.

\begin{figure}
\centering
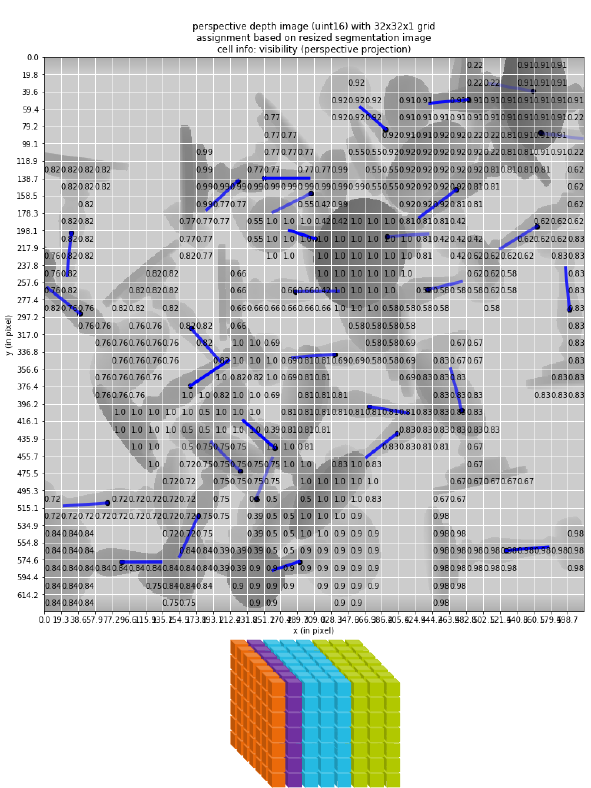
\caption{
(top) Exemplary ground truth sample for the assignment based on resized segmentation images (SI) for the candlestick object from the Sil\'{e}ane dataset~\cite{Bregier.2017}.
(bottom) Resulting output tensor.
%(shape unchanged). # finer grid
}
\label{fig:SI}
\end{figure}

%%%%%%%%%%%%%%%%%%%%%%%%%%%%%%%%%%%%%%%%%%%%%%%%%%%%%%%%%%%%%%%%%%%%%%%%%%%%%%%%
\subsection{Loss Function}
\label{sec:loss_function}
To train the network, the multi-task loss function
% sum over Sx, Sy, Sz, P
\begin{equation}
\mathcal{L}\!=\!
\sum_{i=1}^{S_x}\!
\sum_{j=1}^{S_y}\!
\sum_{k=1}^{S_z}\!
\sum_{l=1}^{P}\!
%\bigg(\!\lambda_1(p_i\!-\!\hat{p}_i)^2+\Big[ \lambda_2(v_i\!-\!\hat{v}_i)^2\!+\!\lambda_3\big(\!\mathcal{L}_\mathrm{pos}\!+\!\lambda_4 \mathcal{L}_\mathrm{ori}\!\big)\!\Big] p_i\!\bigg)
\bigg(\!\lambda_1\mathcal{L}_\mathrm{p}\!+\!\Big[\! \lambda_2\mathcal{L}_\mathrm{v}\!+\!\lambda_3\big(\!\mathcal{L}_\mathrm{pos}\!+\!\lambda_4 \mathcal{L}_\mathrm{ori}\!\big)\!\Big] p_{ijkl}\!\bigg)
\end{equation}
is optimized. The scalars $\lambda_1=0.1$, $\lambda_2=0.25$, $\lambda_3=8v^3$, and $\lambda_4=1$ are used for weighting the different loss terms.

We use a squared L2-loss
%the binary cross-entropy loss
for the presence of an object and the visibility
%channel
to compute $\mathcal{L}_\mathrm{p}$ and $\mathcal{L}_\mathrm{v}$, respectively.
%%%while only backpropagating the loss for the elements that contain ground truth
%%%%by multiplying with $p_i \in \{0,1\}$.
%%by multiplying each channel element-wise with the ground truth probability channel.
%
% position:
% estimate position relative to the grid cell (original, Z)
% estimate the position relative to the image for the variants where the object is assigned to multiple cells (EVE, AP).
Using the squared L2-norm, the positions $x, y, z \in [0,1]$ of the object are estimated relative to the volume element, except for the variants, which make multiple predictions for one object (\mbox{OP-Net} EVE, AP, SI) because the object origin can be located outside of the spatial location. For these variants, the position is estimated relative to an
%the image.
%bigger reference
image reference which is enlarged by the object diameter (diameter of the smallest bounding sphere of the object).
%enlarged image reference, 
This also allows to predict origins outside of the image.
%We do not estimate the position in sensor coordinates to be independent of the scaling.
% angles:
% angles: same loss function as OP-Net
We also use the squared L2-norm for the regression of the Euler angles with bounded values, i.e., $\varphi_1,\varphi_2 \in [0,2\pi)$ and $\varphi_3 \in [0,2\pi/k)$ are mapped to $[0,1)$, where $k \in \mathbb{N}$ represents the order of the cyclic symmetry (similar to vanilla \mbox{OP-Net}~\cite{OP_Net}).
%
%To not waste network capacity and to ease the learning process, we do not enforce a regression of the empty volume elements to zero for all channels, but the channel for the presence of an object origin.
% requires a merging; multiple predictions for the same instance
% see section XX (pose averaging)

%%%%%%%%%%%%%%%%%%%%%%%%%%%%%%%%%%%%%%%%%%%%%%%%%%%%%%%%%%%%%%%%%%%%%%%%%%%%%%%%
\subsection{Pose Averaging}
\label{sec:pose_averaging}
% Average the predictions being close in pose space
% needed for EVE and AP
The variants \mbox{OP-Net} EVE, AP, and SI potentially output multiple predictions per object.
Since these should be close to each other in the 6D pose space and in order not to have many duplicates,
%
%Because these should be located close in 6D pose space and to not have many duplicates,
% identify clusters in 6D space
% For this,
we use unsupervised learning and make use of density-based clustering~\cite{DBSCAN_1,DBSCAN_2}.
%in 6D pose space to identify pose clusters.
To form the final pose predictions, we simply average the resulting pose clusters.
%((predictions being close in pose space).

% weighted averaging?!

% --> averaging needed
% how to balance t and r?!?!

%%%%%%%%%%%%%%%%%%%%%%%%%%%%%%%%%%%%%%%%%%%%%%%%%%%%%%%%%%%%%%%%%%%%%%%%%%%%%%%%
\subsection{Sim-to-Real Transfer}
For a robust transfer of the models from simulation to the real world, we use domain randomization~\cite{DomainRandomization}. The technique to resize the image from the original size to the input size of the neural network (\mbox{$128 \times 128$} pixel) is randomized during training of the model as a kind of noise by randomly selecting a resizing option available in OpenCV. Furthermore, we apply augmentations in a random order and with varying intensity, such as elastic transformations, blurring, and adding noise to the pixels of the image.

% randomize the resizing technique
% elastic transformation(s)
% blurring
% add noise

%%%%%%%%%%%%%%%%%%%%%%%%%%%%%%%%%%%%%%%%%%%%%%%%%%%%%%%%%%%%%%%%%%%%%%%%%%%%%%%%
%\section{Experiments}
\section{Experimental Evaluation}

\begin{figure*}[htbp]
\centering
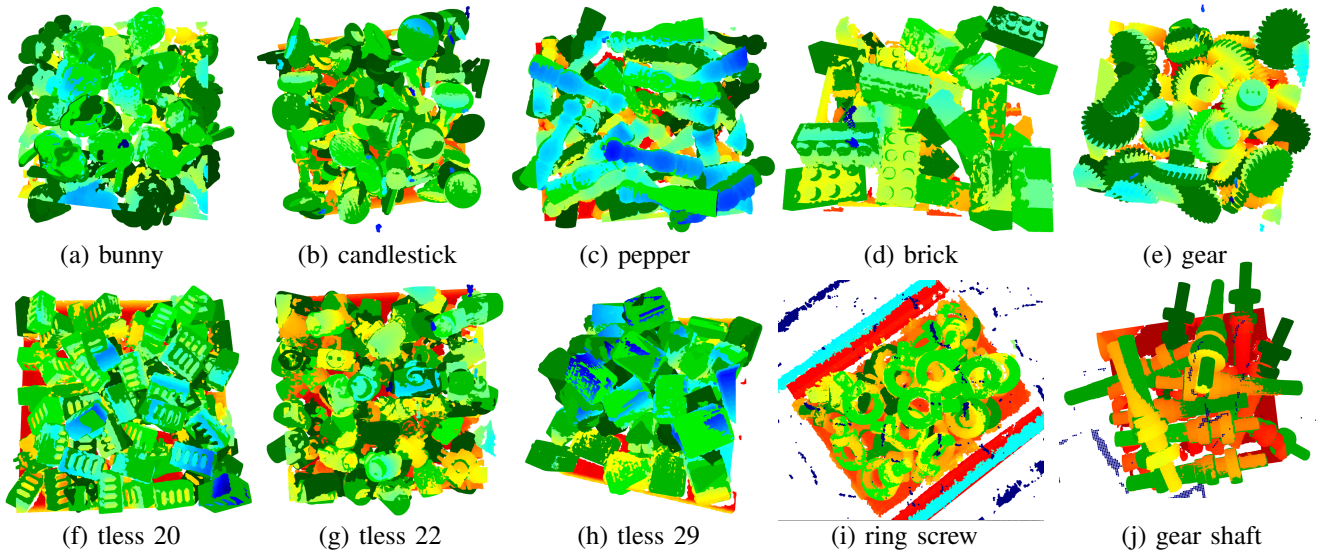
\caption{Qualitative results
% (of our model on noisy real-world data)
on the
% object datasets from the
Sil\'{e}ane~\cite{Bregier.2017}~(a--h) and Fraunhofer IPA~\cite{Fraunhofer_IPA_Bin_Picking_dataset}~(i, j) datasets of \mbox{OP-Net AP} without ICP refinement: 3D point cloud (colored) with pose estimates in green (the brighter, the higher the confidence). Note that the models were trained on synthetic images and annotations only and provide robust results on noisy real-world data.
}
\label{fig:results_datasets}
\end{figure*}

%%%%%%%%%%%%%%%%%%%%%%%%%%%%%%%%%%%%%%%%%%%%%%%%%%%%%%%%%%%%%%%%%%%%%%%%%%%%%%%%
\subsection{Datasets and Evaluation Metric}

We evaluate the performance of our variants of \mbox{OP-Net}
% we evaluate the approaches
on the
% datasets from
Sil\'{e}ane~\cite{Bregier.2017} and Fraunhofer IPA~\cite{Fraunhofer_IPA_Bin_Picking_dataset} datasets. The Fraunhofer IPA dataset is a large-scale dataset that comprises synthetic
%training
data for training deep neural networks and fully pose annotated real-world data for performance evaluation.
We use the synthetic datasets for training our models.
%
% Challenges:
These datasets show scenarios that are typical for bin-picking (see Fig.~\ref{fig:real_world_robot_cell_and_pose_estimates}~(a)) and are very challenging because of a high amount of clutter and heavy occlusions in combination with sensor noise, wrong, or missing depth information.

For evaluation, we use the metric from Br\'{e}gier et al.~\cite{Bregier_Pose_Distance,Bregier.2017}, which considers all possible kinds of object symmetries, can handle scenarios in bulk, requires the recovery of all objects with a visibility of 50\% or higher, and breaks down the performance of a method to a single scalar value (average precision).

%\begin{table}[h]
%\begin{table*}[btp]
\begin{table*}[htp]$ $
\begin{tiny}
%\begin{scriptsize}
\caption{Average precision values of different methods (best results in bold) on the objects from the Sil\'{e}ane~\cite{Bregier.2017} and Fraunhofer IPA~\cite{Fraunhofer_IPA_Bin_Picking_dataset} datasets. For our variants, we did not use ICP refinement. Results marked with * are taken from the ``Object Pose Estimation Challenge for Bin-Picking'' at IROS 2019~\cite{OPE_Challenge_Bin_Picking_IROS}.
}
%\begin{spacing}{0.95}
\begin{center}
\begin{tabular}{l|c|c|c|c|c|c|c|c|c|c|c}
%\hline
object & bunny~\cite{Bregier.2017} & candlestick~\cite{Bregier.2017} & pepper~\cite{Bregier.2017} & brick~\cite{Bregier.2017} & gear~\cite{Bregier.2017} & tless 20~\cite{Bregier.2017} & tless 22~\cite{Bregier.2017} & tless 29~\cite{Bregier.2017} & ring screw~\cite{Fraunhofer_IPA_Bin_Picking_dataset} & gear shaft~\cite{Fraunhofer_IPA_Bin_Picking_dataset} & average \\
object symmetry based on~\cite{Bregier_Pose_Distance,Bregier.2017} & (no proper & (revolution) & (revolution) & (cyclic, & (revolution) & (cyclic, & (no proper & (cyclic, & (cyclic, & (revolution) & \\
& symmetry)  &              &              & $k=2$)  &              & $k=2$)  & symmetry)  & $k=2$) & $k=2$) & & \\
\hline
%%%%%%%%%%
PPF~\cite{PPF,Bregier.2017} & 0.29 & 0.16 & 0.06 & 0.08 & 0.62 & 0.20 & 0.08 & 0.19 & - & - & - \\
PPF PP~\cite{PPF,Bregier.2017} & 0.37 & 0.22 & 0.12 & 0.13 & 0.63 & 0.23 & 0.12 & 0.23 & - & - & - \\
LINEMOD+~\cite{Hinterstoisser_ADD/ADI_LINEMOD+,Bregier.2017} & 0.39 & 0.38 & 0.04 & 0.31 & 0.44 & 0.25 & 0.19 & 0.20 & - & - & - \\
LINEMOD+ PP~\cite{Hinterstoisser_ADD/ADI_LINEMOD+,Bregier.2017} & 0.45 & 0.49 & 0.03 & 0.39 & 0.50 & 0.31 & 0.21 & 0.26 & - & - & - \\
Sock et al.~\cite{Sock} & 0.74 & 0.64 & 0.43 & - & - & - & - & - & - & - & - \\
PPR-Net~\cite{PPR_Net} & 0.82 & 0.91 & 0.80 & - & - & 0.81 & - & - & 0.95* & 0.99* & - \\
PPR-Net with ICP~\cite{PPR_Net} & 0.89 & 0.95 & 0.84 & - & - & 0.85 & - & - & - & - & - \\
%\hline
\mbox{OP-Net} with $\mathcal{L}_\mathrm{ori1}$~\cite{OP_Net} & \textbf{0.92} & 0.94 & 0.98 & 0.41 & \textbf{0.82} & 0.85 & 0.77 & 0.51 & 0.88 & 0.99 & 0.81 \\
%\mbox{OP-Net} with $\mathcal{L}_\mathrm{ori1}$ and PP~\cite{OP_Net} & \textbf{0.94} & \textbf{0.97} & \textbf{0.98} & 0.42 & \textbf{0.84} & \textbf{0.88} & \textbf{0.86} & \textbf{0.58} & 0.93 & 0.99 & - \\
\mbox{OP-Net} with $\mathcal{L}_\mathrm{ori2}$~\cite{OP_Net} & 0.74 & 0.95 & 0.92 & \textbf{0.79} & 0.58 & 0.56 & 0.53 & 0.36 & 0.73 & \textbf{1.0} & 0.72 \\
%\mbox{OP-Net} with $\mathcal{L}_\mathrm{ori2}$ and PP~\cite{OP_Net} & 0.76 & 0.96 & 0.93 & \textbf{0.80} & 0.60 & 0.58 & 0.55 & 0.39 & 0.75 & \textbf{1.0} & - \\
\hline
%OP-Net AD (ours) & 0.xx & 0.xx & 0.xx & 0.xx & 0.xx & 0.xx & 0.xx & 0.xx & 0.xx & 0.xx \\
%
%OP-Net~\cite{OP_Net} with $S_x=S_y=32$ & 0.85 & 0.95 & 0.97 & 0.42 & 0.79 & 0.87 & 0.83 & 0.44 & 0.82 & 0.97 & 0.79 \\
%
OP-Net EVE (ours) & \textbf{0.92} & 0.94 & 0.97 & 0.45 & 0.77 & \textbf{0.88} & 0.81 & \textbf{0.56} & 0.94 & 0.98 & 0.82 \\
OP-Net AP (ours) & \textbf{0.92} & \textbf{0.98} & \textbf{0.99} & 0.45 & \textbf{0.82} & 0.87 & \textbf{0.84} & \textbf{0.56} & \textbf{0.96} & 0.99 & \textbf{0.84} \\
OP-Net Z (ours) & 0.85 & 0.92 & 0.95 & 0.41 & 0.72 & 0.84 & \textbf{0.84} & 0.43 & 0.92 & 0.95 & 0.78 \\
OP-Net MP (ours) & 0.92 & 0.82 & 0.91 & 0.32 & 0.80 & 0.70 & 0.66 & \textbf{0.56} & 0.84 & 0.95 & 0.75 \\
OP-Net SI (ours) & 0.30 & 0.87 & 0.97 & 0.36 & 0.37 & 0.46 & 0.51 & 0.36 & 0.56 & 0.88 & 0.56 \\
%\hline
\end{tabular}
\end{center}
%\end{spacing}
\label{table:results_on_datasets}
\end{tiny}
%\end{scriptsize}
%\vspace{-3mm}
\end{table*}

%%%%%%%%%%%%%%%%%%%%%%%%%%%%%%%%%%%%%%%%%%%%%%%%%%%%%%%%%%%%%%%%%%%%%%%%%%%%%%%%
\subsection{Setup}

% neural network:
We use a fully convolutional network architecture to learn the mapping from normalized input depth images in perspective projection to the output tensors, where $S_x$ and $S_y$ correspond to the size of the feature map of the last layer of the neural network. This method is a single shot approach because only one forward pass of the neural network is needed instead of multiple evaluations at different locations.
%For best comparability, we use a DenseNet-BC~\cite{DenseNet,DenseNet_JournalPaper} with 40 layers and a growth rate of 50, which indicates the number of feature maps being added per layer, analogously to \mbox{OP-Net}~\cite{OP_Net}.

% use the same network architecture, etc. as OP-Net
Analogously to \mbox{OP-Net}~\cite{OP_Net},
%(for best comparability),
we use a DenseNet-BC~\cite{DenseNet,DenseNet_JournalPaper} with 40 layers and a growth rate of 50, which indicates the number of feature maps being added per layer.
%For a fair comparison,
For best comparability, we also use the same loss function for the regression of the orientation (see Sec.~\ref{sec:loss_function}) and an input resolution of \mbox{$128 \times 128$} pixel. For the variants \mbox{OP-Net} EVE and AP, we use \mbox{$S_x=S_y=16$}, similar to vanilla \mbox{OP-Net} for best comparability.

%%%%%%%%%%%%%%%%%%%%%%%%%%%%%%%%%%%%%%%%%%%%%%%%%%%%%%%%%%%%%%%%%%%%%%%%%%%%%%%%
\subsection{Results}
Our variants provide robust pose estimates although they are fully trained on synthetic images and annotations. The test datasets were recorded with different 3D sensors, i.e., providing robust pose estimates is independent of the actual 3D sensor technology being used.
% At test time on real-world data,
Pixels with missing depth information are bilinearly interpolated
%interpolated (bilinear)
before being fed into the neural network.
Table~\ref{table:results_on_datasets} reports the performance of the different variants in terms of average precision, where all values of \mbox{OP-Net} and its variants are without ICP refinement.
%
%%%%%%%%%%%
% results %
%%%%%%%%%%%
% AP and EVE to address all 3 problems/limitations?
%We also show that simply using a fine grid does improve the performance.
%$S_x=S_y=32$ and ($S_z=1$, $P=1$) does not solve the problem
%We demonstrate that simply using a finer discretization for \mbox{OP-Net}~\cite{OP_Net}
%specifying a smaller grid
%with $S_x=S_y=32$, does not or for some datasets only slightly improve the performance.
%
Our variants outperform other classical and learning-based approaches on multiple
%/all?
datasets. Fig.~\ref{fig:results_datasets} visualizes qualitative results of \mbox{OP-Net AP} on the different datasets.

% OP-Net EVE
The variant \mbox{OP-Net EVE} can give slightly better results because it also outputs useful pose information for challenging border cases.
%
% OP-Net AP
\mbox{OP-Net AP}
% AP does not hurt performance for objects where AP are not relevant/needed?! --> would be nice result
%... can handle even heavier occlusions!!!
% ... improved the performance for on most datasets and did not harm the performance for the other/remaining objects.
performs significantly better than other state-of-the-art approaches because it misses fewer objects in the parameterization and addresses all major limitations in Fig.~\ref{fig:limitations}.
%
% OP-Net Z
%The 3D discretization demonstrates high AP values but does not outperform OP-Net.
\mbox{OP-Net} Z and MP also provide good results but not consistently along all datasets. A key limitation is the very high-dimensional output tensor.
%
% OP-Net SI
For \mbox{OP-Net SI}, we encountered the transfer from synthetic to real-world data as a key limitation due to the dense parameterization of the output.

% Time advantage:
The \mbox{OP-Net} variants are approaches which require a single forward pass of the network only (single shot) with 13 or 17~ms per depth image for the forward pass on a Nvidia Tesla V100 or GTX 1080 Ti, respectively.
With 200~ms
%(Nvidia Titan
for the forward pass (GTX 1060) plus the time for the additional clustering and averaging of many datapoints, PPR-Net~\cite{PPR_Net} is much slower than OP-Net and its variants.

% faster than bp3 on the same hardware...

%to be able to benefit form the high speed (13 ms or 17 ms per depth image for the forward pass on a Nvidia Tesla V100 or GTX 1080Ti, respectively) and real-time capabilities of single shot approaches, we investigate in different novel parameterizations of the output to increase the performance.

% OP-Net AD
%... OP-Net AD performs worse, also with larger input resolution.
%while also getting a larger and therefore more detailed input depth image.

%%%%%%%%%%%%%%%%%%%%%%%%%%%%%%%%%%%%%%%%%%%%%%%%%%%%%%%%%%%%%%%%%%%%%%%%%%%%%%%%
\section{Conclusions}
In this work, we propose different novel parameterizations of the output for single shot approaches to 6D object pose estimation. Our experiments demonstrate that extending the spatial locations
% can boost the performance and especially
or adding additional
%(reference)
points to the object significantly
%boosts/
improves the performance in terms of average precision, allows overcoming the limitations of previous works, and improves the performance of single shot approaches.
Our models demonstrate state-of-the-art performance on various
%(/all?)
public datasets and can be used for real-world robotic grasping tasks without ICP refinement.
% ICP recommended for precise placement

%%%In future work, we plan to study the influence of the projection type (orthogonal vs. perspective), different loss functions (e.g., for the regression of the orientation), and different parameterizations for the orientation (quaternions, axis-angle, rotation matrix).

\addtolength{\textheight}{-7cm}  % This command serves to balance the column lengths

%%%%%%%%%%%%%%%%%%%%%%%%%%%%%%%%%%%%%%%%%%%%%%%%%%%%%%%%%%%%%%%%%%%%%%%%%%%%%%%%

%%%%%%%%%%%%%%%%%%%%%%%%%%%%%%%%%%%%%%%%%%%%%%%%%%%%%%%%%%%%%%%%%%%%%%%%%%%%%%%%

%%%%%%%%%%%%%%%%%%%%%%%%%%%%%%%%%%%%%%%%%%%%%%%%%%%%%%%%%%%%%%%%%%%%%%%%%%%%%%%%
\section*{Acknowledgment}
This work was partially supported by the Federal Ministry of Education and Research (Deep Picking -- Grant No. 01IS20005C) and the
State Ministry of Baden-W\"urttemberg for Economic Affairs, Labour and Housing Construction
%Ministry of Economic Affairs of the state Baden-W\"urttemberg
(Center for Cognitive Robotics –- Grant No. 017-180004 and Center for Cyber Cognitive Intelligence (CCI) -- Grant No. 017-192996).
%We would like to thank our colleagues for helpful discussions and comments.

%%%%%%%%%%%%%%%%%%%%%%%%%%%%%%%%%%%%%%%%%%%%%%%%%%%%%%%%%%%%%%%%%%%%%%%%%%%%%%%%

\bibliographystyle{IEEEtran}
\bibliography{IEEEabrv,references}

\end{document}

%% file: figure_cover.pdf_tex
%% Accompanies image file 'figure_cover.pdf' (pdf, eps, ps)
\begingroup%
  \makeatletter%
  \providecommand\color[2][]{%
    \errmessage{(Inkscape) Color is used for the text in Inkscape, but the package 'color.sty' is not loaded}%
    \renewcommand\color[2][]{}%
  }%
  \providecommand\transparent[1]{%
    \errmessage{(Inkscape) Transparency is used (non-zero) for the text in Inkscape, but the package 'transparent.sty' is not loaded}%
    \renewcommand\transparent[1]{}%
  }%
  \providecommand\rotatebox[2]{#2}%
  \newcommand*\fsize{\dimexpr\f@size pt\relax}%
  \newcommand*\lineheight[1]{\fontsize{\fsize}{#1\fsize}\selectfont}%
  \ifx\svgwidth\undefined%
    \setlength{\unitlength}{226.77165354bp}%
    \ifx\svgscale\undefined%
      \relax%
    \else%
      \setlength{\unitlength}{\unitlength * \real{\svgscale}}%
    \fi%
  \else%
    \setlength{\unitlength}{\svgwidth}%
  \fi%
  \global\let\svgwidth\undefined%
  \global\let\svgscale\undefined%
  \makeatother%
  \begin{picture}(1,1.4375)%
    \lineheight{1}%
    \setlength\tabcolsep{0pt}%
    \put(0,0){\includegraphics[width=\unitlength,page=1]{figure_cover.pdf}}%
    \put(0.47578931,0.70953693){\color[rgb]{0,0,0}\makebox(0,0)[lt]{\lineheight{1.25}\smash{\begin{tabular}[t]{l}(a)\end{tabular}}}}%
    \put(0.47578931,-0.02386818){\color[rgb]{0,0,0}\makebox(0,0)[lt]{\lineheight{1.25}\smash{\begin{tabular}[t]{l}(b)\end{tabular}}}}%
  \end{picture}%
\endgroup%

%% file: figure_limitations.pdf_tex
%% Accompanies image file 'figure_limitations.pdf' (pdf, eps, ps)
\begingroup%
  \makeatletter%
  \providecommand\color[2][]{%
    \errmessage{(Inkscape) Color is used for the text in Inkscape, but the package 'color.sty' is not loaded}%
    \renewcommand\color[2][]{}%
  }%
  \providecommand\transparent[1]{%
    \errmessage{(Inkscape) Transparency is used (non-zero) for the text in Inkscape, but the package 'transparent.sty' is not loaded}%
    \renewcommand\transparent[1]{}%
  }%
  \providecommand\rotatebox[2]{#2}%
  \newcommand*\fsize{\dimexpr\f@size pt\relax}%
  \newcommand*\lineheight[1]{\fontsize{\fsize}{#1\fsize}\selectfont}%
  \ifx\svgwidth\undefined%
    \setlength{\unitlength}{481.88976378bp}%
    \ifx\svgscale\undefined%
      \relax%
    \else%
      \setlength{\unitlength}{\unitlength * \real{\svgscale}}%
    \fi%
  \else%
    \setlength{\unitlength}{\svgwidth}%
  \fi%
  \global\let\svgwidth\undefined%
  \global\let\svgscale\undefined%
  \makeatother%
  \begin{picture}(1,0.33529412)%
    \lineheight{1}%
    \setlength\tabcolsep{0pt}%
    \put(0,0){\includegraphics[width=\unitlength,page=1]{figure_limitations.pdf}}%
    \put(0.13999087,0.0240418){\color[rgb]{0,0,0}\makebox(0,0)[lt]{\begin{minipage}{0.06569066\unitlength}\raggedright (a)\end{minipage}}}%
    \put(0.52692549,0.02404345){\color[rgb]{0,0,0}\makebox(0,0)[lt]{\begin{minipage}{0.06569066\unitlength}\raggedright (b)\end{minipage}}}%
    \put(0.85797631,0.02404345){\color[rgb]{0,0,0}\makebox(0,0)[lt]{\begin{minipage}{0.06569066\unitlength}\raggedright (c)\end{minipage}}}%
    \put(0,0){\includegraphics[width=\unitlength,page=2]{figure_limitations.pdf}}%
  \end{picture}%
\endgroup%

%% file: EVE.pdf_tex
%% Accompanies image file 'EVE.pdf' (pdf, eps, ps)
\begingroup%
  \makeatletter%
  \providecommand\color[2][]{%
    \errmessage{(Inkscape) Color is used for the text in Inkscape, but the package 'color.sty' is not loaded}%
    \renewcommand\color[2][]{}%
  }%
  \providecommand\transparent[1]{%
    \errmessage{(Inkscape) Transparency is used (non-zero) for the text in Inkscape, but the package 'transparent.sty' is not loaded}%
    \renewcommand\transparent[1]{}%
  }%
  \providecommand\rotatebox[2]{#2}%
  \newcommand*\fsize{\dimexpr\f@size pt\relax}%
  \newcommand*\lineheight[1]{\fontsize{\fsize}{#1\fsize}\selectfont}%
  \ifx\svgwidth\undefined%
    \setlength{\unitlength}{189.92125984bp}%
    \ifx\svgscale\undefined%
      \relax%
    \else%
      \setlength{\unitlength}{\unitlength * \real{\svgscale}}%
    \fi%
  \else%
    \setlength{\unitlength}{\svgwidth}%
  \fi%
  \global\let\svgwidth\undefined%
  \global\let\svgscale\undefined%
  \makeatother%
  \begin{picture}(1,1.19402985)%
    \lineheight{1}%
    \setlength\tabcolsep{0pt}%
    \put(0,0){\includegraphics[width=\unitlength,page=1]{EVE.pdf}}%
  \end{picture}%
\endgroup%

%% file: AP.pdf_tex
%% Accompanies image file 'AP.pdf' (pdf, eps, ps)
\begingroup%
  \makeatletter%
  \providecommand\color[2][]{%
    \errmessage{(Inkscape) Color is used for the text in Inkscape, but the package 'color.sty' is not loaded}%
    \renewcommand\color[2][]{}%
  }%
  \providecommand\transparent[1]{%
    \errmessage{(Inkscape) Transparency is used (non-zero) for the text in Inkscape, but the package 'transparent.sty' is not loaded}%
    \renewcommand\transparent[1]{}%
  }%
  \providecommand\rotatebox[2]{#2}%
  \newcommand*\fsize{\dimexpr\f@size pt\relax}%
  \newcommand*\lineheight[1]{\fontsize{\fsize}{#1\fsize}\selectfont}%
  \ifx\svgwidth\undefined%
    \setlength{\unitlength}{481.88976378bp}%
    \ifx\svgscale\undefined%
      \relax%
    \else%
      \setlength{\unitlength}{\unitlength * \real{\svgscale}}%
    \fi%
  \else%
    \setlength{\unitlength}{\svgwidth}%
  \fi%
  \global\let\svgwidth\undefined%
  \global\let\svgscale\undefined%
  \makeatother%
  \begin{picture}(1,0.21764706)%
    \lineheight{1}%
    \setlength\tabcolsep{0pt}%
    \put(0,0){\includegraphics[width=\unitlength,page=1]{AP.pdf}}%
    \put(0.19460614,0.01642412){\color[rgb]{0,0,0}\makebox(0,0)[lt]{\begin{minipage}{0.03371836\unitlength}\centering (a)\end{minipage}}}%
    \put(0.63906165,0.01642412){\color[rgb]{0,0,0}\makebox(0,0)[lt]{\begin{minipage}{0.03371836\unitlength}\centering (b)\end{minipage}}}%
    \put(0,0){\includegraphics[width=\unitlength,page=2]{AP.pdf}}%
    \put(0.9315482,0.01642124){\color[rgb]{0,0,0}\makebox(0,0)[lt]{\begin{minipage}{0.03371836\unitlength}\centering (c)\end{minipage}}}%
  \end{picture}%
\endgroup%

%% file: Z.pdf_tex
%% Accompanies image file 'Z.pdf' (pdf, eps, ps)
\begingroup%
  \makeatletter%
  \providecommand\color[2][]{%
    \errmessage{(Inkscape) Color is used for the text in Inkscape, but the package 'color.sty' is not loaded}%
    \renewcommand\color[2][]{}%
  }%
  \providecommand\transparent[1]{%
    \errmessage{(Inkscape) Transparency is used (non-zero) for the text in Inkscape, but the package 'transparent.sty' is not loaded}%
    \renewcommand\transparent[1]{}%
  }%
  \providecommand\rotatebox[2]{#2}%
  \newcommand*\fsize{\dimexpr\f@size pt\relax}%
  \newcommand*\lineheight[1]{\fontsize{\fsize}{#1\fsize}\selectfont}%
  \ifx\svgwidth\undefined%
    \setlength{\unitlength}{170.07874016bp}%
    \ifx\svgscale\undefined%
      \relax%
    \else%
      \setlength{\unitlength}{\unitlength * \real{\svgscale}}%
    \fi%
  \else%
    \setlength{\unitlength}{\svgwidth}%
  \fi%
  \global\let\svgwidth\undefined%
  \global\let\svgscale\undefined%
  \makeatother%
  \begin{picture}(1,1)%
    \lineheight{1}%
    \setlength\tabcolsep{0pt}%
    \put(0,0){\includegraphics[width=\unitlength,page=1]{Z.pdf}}%
    \put(1.67740452,0.76197646){\color[rgb]{0,0,0}\makebox(0,0)[lt]{\begin{minipage}{0.4976686\unitlength}\raggedright \end{minipage}}}%
    \put(1.69945313,0.6611828){\color[rgb]{0,0,0}\makebox(0,0)[lt]{\begin{minipage}{0.29293169\unitlength}\raggedright \end{minipage}}}%
    \put(0.37252308,0.68856076){\color[rgb]{0,0,0}\makebox(0,0)[lt]{\begin{minipage}{0.33408692\unitlength}\raggedright \end{minipage}}}%
    \put(0.29456945,0.68633351){\color[rgb]{0,0,0}\makebox(0,0)[lt]{\begin{minipage}{0.19599767\unitlength}\raggedright \end{minipage}}}%
    \put(0.51363154,0.66475681){\color[rgb]{0,0,0}\makebox(0,0)[lt]{\begin{minipage}{0.57771945\unitlength}\centering near clipping plane\end{minipage}}}%
    \put(0.64837741,0.34184447){\color[rgb]{0,0,0}\makebox(0,0)[lt]{\begin{minipage}{0.57771945\unitlength}\centering far clipping plane\end{minipage}}}%
    \put(1.42053974,0.91387829){\color[rgb]{0,0,0}\makebox(0,0)[lt]{\begin{minipage}{0.4976686\unitlength}\raggedright \end{minipage}}}%
    \put(1.44258835,0.81308463){\color[rgb]{0,0,0}\makebox(0,0)[lt]{\begin{minipage}{0.29293169\unitlength}\raggedright \end{minipage}}}%
    \put(0.1156583,0.84046259){\color[rgb]{0,0,0}\makebox(0,0)[lt]{\begin{minipage}{0.33408692\unitlength}\raggedright \end{minipage}}}%
    \put(0.03770467,0.83823534){\color[rgb]{0,0,0}\makebox(0,0)[lt]{\begin{minipage}{0.19599767\unitlength}\raggedright \end{minipage}}}%
    \put(0.44654997,0.96920256){\color[rgb]{0,0,0}\makebox(0,0)[lt]{\begin{minipage}{0.21803383\unitlength}\centering sensor\end{minipage}}}%
    \put(0,0){\includegraphics[width=\unitlength,page=2]{Z.pdf}}%
  \end{picture}%
\endgroup%

%% file: MP.pdf_tex
%% Accompanies image file 'MP.pdf' (pdf, eps, ps)
\begingroup%
  \makeatletter%
  \providecommand\color[2][]{%
    \errmessage{(Inkscape) Color is used for the text in Inkscape, but the package 'color.sty' is not loaded}%
    \renewcommand\color[2][]{}%
  }%
  \providecommand\transparent[1]{%
    \errmessage{(Inkscape) Transparency is used (non-zero) for the text in Inkscape, but the package 'transparent.sty' is not loaded}%
    \renewcommand\transparent[1]{}%
  }%
  \providecommand\rotatebox[2]{#2}%
  \newcommand*\fsize{\dimexpr\f@size pt\relax}%
  \newcommand*\lineheight[1]{\fontsize{\fsize}{#1\fsize}\selectfont}%
  \ifx\svgwidth\undefined%
    \setlength{\unitlength}{240.94488189bp}%
    \ifx\svgscale\undefined%
      \relax%
    \else%
      \setlength{\unitlength}{\unitlength * \real{\svgscale}}%
    \fi%
  \else%
    \setlength{\unitlength}{\svgwidth}%
  \fi%
  \global\let\svgwidth\undefined%
  \global\let\svgscale\undefined%
  \makeatother%
  \begin{picture}(1,0.47058824)%
    \lineheight{1}%
    \setlength\tabcolsep{0pt}%
    \put(0,0){\includegraphics[width=\unitlength,page=1]{MP.pdf}}%
  \end{picture}%
\endgroup%

%% file: SI.pdf_tex
%% Accompanies image file 'SI.pdf' (pdf, eps, ps)
\begingroup%
  \makeatletter%
  \providecommand\color[2][]{%
    \errmessage{(Inkscape) Color is used for the text in Inkscape, but the package 'color.sty' is not loaded}%
    \renewcommand\color[2][]{}%
  }%
  \providecommand\transparent[1]{%
    \errmessage{(Inkscape) Transparency is used (non-zero) for the text in Inkscape, but the package 'transparent.sty' is not loaded}%
    \renewcommand\transparent[1]{}%
  }%
  \providecommand\rotatebox[2]{#2}%
  \newcommand*\fsize{\dimexpr\f@size pt\relax}%
  \newcommand*\lineheight[1]{\fontsize{\fsize}{#1\fsize}\selectfont}%
  \ifx\svgwidth\undefined%
    \setlength{\unitlength}{170.07874016bp}%
    \ifx\svgscale\undefined%
      \relax%
    \else%
      \setlength{\unitlength}{\unitlength * \real{\svgscale}}%
    \fi%
  \else%
    \setlength{\unitlength}{\svgwidth}%
  \fi%
  \global\let\svgwidth\undefined%
  \global\let\svgscale\undefined%
  \makeatother%
  \begin{picture}(1,1.33333333)%
    \lineheight{1}%
    \setlength\tabcolsep{0pt}%
    \put(0,0){\includegraphics[width=\unitlength,page=1]{SI.pdf}}%
  \end{picture}%
\endgroup%

%% file: results.pdf_tex
%% Accompanies image file 'results.pdf' (pdf, eps, ps)
\begingroup%
  \makeatletter%
  \providecommand\color[2][]{%
    \errmessage{(Inkscape) Color is used for the text in Inkscape, but the package 'color.sty' is not loaded}%
    \renewcommand\color[2][]{}%
  }%
  \providecommand\transparent[1]{%
    \errmessage{(Inkscape) Transparency is used (non-zero) for the text in Inkscape, but the package 'transparent.sty' is not loaded}%
    \renewcommand\transparent[1]{}%
  }%
  \providecommand\rotatebox[2]{#2}%
  \newcommand*\fsize{\dimexpr\f@size pt\relax}%
  \newcommand*\lineheight[1]{\fontsize{\fsize}{#1\fsize}\selectfont}%
  \ifx\svgwidth\undefined%
    \setlength{\unitlength}{496.06299213bp}%
    \ifx\svgscale\undefined%
      \relax%
    \else%
      \setlength{\unitlength}{\unitlength * \real{\svgscale}}%
    \fi%
  \else%
    \setlength{\unitlength}{\svgwidth}%
  \fi%
  \global\let\svgwidth\undefined%
  \global\let\svgscale\undefined%
  \makeatother%
  \begin{picture}(1,0.42857143)%
    \lineheight{1}%
    \setlength\tabcolsep{0pt}%
    \put(0.04587891,0.21795881){\color[rgb]{0,0,0}\makebox(0,0)[lt]{\lineheight{1.25}\smash{\begin{tabular}[t]{l}(a) bunny\end{tabular}}}}%
    \put(0.22646465,0.21795881){\color[rgb]{0,0,0}\makebox(0,0)[lt]{\lineheight{1.25}\smash{\begin{tabular}[t]{l}(b) candlestick\end{tabular}}}}%
    \put(0.44028769,0.21795881){\color[rgb]{0,0,0}\makebox(0,0)[lt]{\lineheight{1.25}\smash{\begin{tabular}[t]{l}(c) pepper\end{tabular}}}}%
    \put(0.65641,0.21795881){\color[rgb]{0,0,0}\makebox(0,0)[lt]{\lineheight{1.25}\smash{\begin{tabular}[t]{l}(d) brick\end{tabular}}}}%
    \put(0.04723561,0.00392456){\color[rgb]{0,0,0}\makebox(0,0)[lt]{\lineheight{1.25}\smash{\begin{tabular}[t]{l}(f) tless 20\end{tabular}}}}%
    \put(0.24245246,0.00402937){\color[rgb]{0,0,0}\makebox(0,0)[lt]{\lineheight{1.25}\smash{\begin{tabular}[t]{l}(g) tless 22\end{tabular}}}}%
    \put(0.43871822,0.00392456){\color[rgb]{0,0,0}\makebox(0,0)[lt]{\lineheight{1.25}\smash{\begin{tabular}[t]{l}(h) tless 29\end{tabular}}}}%
    \put(0.63143079,0.00402937){\color[rgb]{0,0,0}\makebox(0,0)[lt]{\lineheight{1.25}\smash{\begin{tabular}[t]{l}(i) ring screw\end{tabular}}}}%
    \put(0.85215114,0.00402937){\color[rgb]{0,0,0}\makebox(0,0)[lt]{\lineheight{1.25}\smash{\begin{tabular}[t]{l}(j) gear shaft\end{tabular}}}}%
    \put(0,0){\includegraphics[width=\unitlength,page=1]{results.pdf}}%
    \put(0.8683531,0.21806362){\color[rgb]{0,0,0}\makebox(0,0)[lt]{\lineheight{1.25}\smash{\begin{tabular}[t]{l}(e) gear\end{tabular}}}}%
  \end{picture}%
\endgroup%